\documentclass[conference]{IEEEtran}
\IEEEoverridecommandlockouts
\usepackage{cite}
\usepackage{amsmath,amssymb,amsfonts}
\usepackage{algorithmic}
\usepackage{graphicx}
\usepackage{textcomp}
\usepackage{xcolor}
\def\BibTeX{{\rm B\kern-.05em{\sc i\kern-.025em b}\kern-.08em
    T\kern-.1667em\lower.7ex\hbox{E}\kern-.125emX}}

\usepackage{algorithm}
\usepackage{algorithmic}
\usepackage{amsmath}
\usepackage{booktabs}
\usepackage{amsfonts}
\usepackage{multirow}
\usepackage{hyperref}

\begin{document}

\title{Enhancing Large Multimodal Models with Adaptive Sparsity and KV Cache Compression}

    \author{
    \IEEEauthorblockN{Te Zhang$^{1}$\textsuperscript{*}\footnotemark,Yuheng Li$^{2}$, Junxiang Wang$^{3}$, Lujun Li$^{4}$ }
    \IEEEauthorblockA{
    \\$^{1}$University of Michigan, $^{2}$ Johns Hopkins University, $^{3}$Central South university, $^{4}$ HKGAI
\\\texttt{$^{1}$zte@umich.edu}, \texttt{$^{2}$yli623@jh.edu}, \texttt{$^{3}$blp2024\_cmt@163.com}, \texttt{$^{4}$lujunli@hkgai.org}
  }}

% \thanks{First and corresponding author: Te Zhang. Yuheng Li with equal contributions in rebuttal and camera-ready preparation.}
\maketitle
% \footnotetext[*]{Corresponding author: email@example.com}
\renewcommand{\thefootnote}{*}
\footnotetext{First and corresponding author: Te Zhang. Yuheng Li contributed equally in the rebuttal and camera-ready preparation.}
\renewcommand{\thefootnote}{\arabic{footnote}}
\begin{abstract}
Large multimodal models (LMMs) have advanced significantly by integrating visual encoders with extensive language models, enabling robust reasoning capabilities. However, compressing LMMs for deployment on edge devices remains a critical challenge. In this work, we propose an adaptive search algorithm that optimizes sparsity and KV cache compression to enhance LMM efficiency. Utilizing the Tree-structured Parzen Estimator, our method dynamically adjusts pruning ratios and KV cache quantization bandwidth across different LMM layers, using model performance as the optimization objective. This approach uniquely combines pruning with key-value cache quantization and incorporates a fast pruning technique that eliminates the need for additional fine-tuning or weight adjustments, achieving efficient compression without compromising accuracy. Comprehensive evaluations on benchmark datasets, including LLaVA-1.5 7B and 13B, demonstrate our method's superiority over state-of-the-art techniques such as SparseGPT and Wanda across various compression levels. Notably, our framework's automatic allocation of KV cache compression resources sets a new standard in LMM optimization, delivering memory efficiency without sacrificing much performance. Code is available at \url{https://github.com/tezhang65/optspa.git}
\end{abstract}

\section{Introduction}
\label{sec:intro}

Recent large multimodal models (LMMs) have achieved state-of-the-art performance in vision-language tasks such as visual question answering and reasoning \cite{jiang2023mistral, touvron2023llama}. These models leverage the reasoning capabilities of large language models (LLMs) and utilize visual encoders like CLIP-ViT to embed image patches as visual tokens \cite{liu2023llava}. However, the high computational and GPU memory requirements have driven the trend towards lightweight LMMs for edge deployment, exemplified by mini versions like MiniGPT-4 \cite{zhu2023minigpt}.
\begin{figure}[ht]
\begin{center}
\centerline{\includegraphics[width=1.0\linewidth]{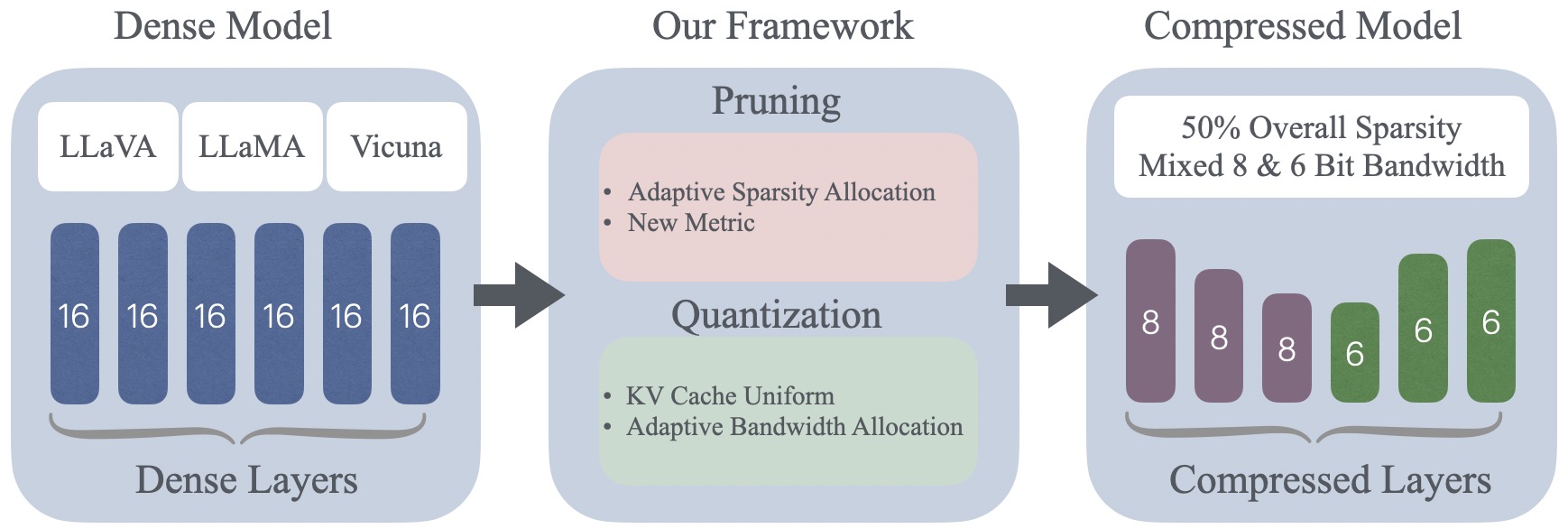}}
  \caption{A brief schematic overview of our framework. We utilize the Tree-structured Parzen estimator to search  pruning ratios and KV Cache quantization bandwidth across different layers. The length of the vertical rectangles indicates the sparsity ratio, the longer the denser. Colors for KV Cache quantization bandwidth: Blue = 16 bit, Purple = 8 bit and Green = 6 bit.
}
  \label{fig:main}
\end{center}
\end{figure}

To reduce computational costs while preserving performance, compression techniques such as distillation~\cite{li2024detkds,li2024kd,Dong2023diswot,li2023automated,li2022norm,li2022self} and Architecture Search~\cite{li2024attnzero,li2024auto} are employed. We conducted extensive experiments to evaluate the impact of these techniques on LMM performance and found that allocating different compression levels across model layers enhances results. This insight inspired the development of a novel framework that integrates pruning and quantization with an optimization search algorithm, intelligently assigning layer-wise hyperparameters (e.g., sparsity ratio and bandwidth) to achieve optimal performance at target compression levels. Additionally, our search results also reveal key insights into the relationship between compression levels and layer depths.

Our framework introduces an automatic layer-wise sparsity allocation search algorithm, and an advanced pruning metric incorporating L2 normalization and logarithmic scaling. These methods enhance model performance compared to uniform sparsity approaches \cite{li2024dis, li2024als, dong2024pruner}. For Key-Value (KV) cache compression, we utilize uniform asymmetric quantization and our search algorithm to optimize quantization bandwidths across different layers, balancing compression efficiency with performance. 

Our contributions can be summarized as follows:
\begin{itemize}
    \item To our knowledge, our framework is the first to integrate both pruning and quantization methods and benchmark them on large multimodal models. Our experiments provide insights into optimal allocation of sparsity ratios and KV cache quantization bandwidths across different layer depths, guiding future compression applications.
    \item We propose a layer-wise allocation method using the TPE algorithm to efficiently search for optimized parameters (e.g., sparsity ratio and bandwidth) in compression techniques. The search results can be served as "compression profiles" for models with varying numbers of layers. Additionally, we introduce an advanced pruning metric that captures weights and activations in both row and column dimensions, enabling accurate ranking of unit importance.
    \item We conducted extensive experiments on benchmarks like VQAv2, SQA, POPE, TextVQA, and Wikitext perplexity. Our pruning and quantization methods demonstrate superior performance in maintaining model quality under various sparsity ratios and quantization bandwidths.
\end{itemize}

\section{Related Works}

\subsection{Large Multimodal Models}

Large Language Models (LLMs) like GPT-4 \cite{gpt4}, LLaMA \cite{touvron2023llama}, have shown impressive reasoning in text tasks. Large Multimodal Models (LMMs) \cite{liu2023llava, zhu2023minigpt, yin2023survey, zhang2024mm} extend these capabilities to images by combining a vision encoder with a pretrained LLM to generate text responses from images and questions. LLaVA uses a simple Multilayer Perceptron (MLP) projector to align image and text information, achieving state-of-the-art performance on over 10 benchmarks. 

\subsection{Sparsity and Pruning}

Pruning reduces the size of the neural network by removing less important weights, traditionally requiring retraining. However, retraining large LLMs is computationally expensive. Recent post-training pruning methods, such as SparseGPT \cite{Frantar2023SparseGPTML} and Wanda \cite{Sun2023ASA_wanda}, avoid retraining, enabling fast pruning with state-of-the-art performance at high sparsity ratios. Other works like LLM-Pruner \cite{Ma2023LLMPrunerOT} propose structured and unstructured pruning to reduce model size and computation. The core of pruning methods is the design of a metric to evaluate the importance of weight. Since layers at different depths respond differently to sparsity, sparsity allocation should be adaptive rather than uniform.

\subsection{KV Cache Compression}

The Key-Value (KV) cache in LLMs stores intermediate results during inference but demands substantial memory, especially with long sequences or large batches \cite{dong2023emq, dong2024stbllm}, making deployment on resource-constrained devices challenging. Compressing the KV cache is essential to reduce memory usage. Quantization reduces matrix precision to lower bandwidth representations and is effective for KV cache compression.  We focus on uniform asymmetric quantization \cite{Jacob2018Quantization}, a fast method with minimal overhead. Our goal is to investigate optimal bit bandwidth allocation across layers to optimize performance without degrading the model's ability.

\section{Methodology}

\subsection{Pruning and Sparsity Allocation}

\noindent \textbf{Problem Definition:} Post-training pruning reduces model size by setting less important weights in each layer's weight matrix \( \mathbf{W} \) to zero, using a 0/1 mask \( \mathbf{M} \) based on a pruning metric \( \mathbf{S} \). An effective metric minimizes the reconstruction loss:

\begin{equation}
\min_{\mathbf{M}_{l}} \| \mathbf{W}_{l}\mathbf{X}_{l} - (\mathbf{M}_{l} \odot \mathbf{W}_{l}) \mathbf{X}_{l} \|_2^2
\label{eq:recon_loss}
\end{equation}

where \( \mathbf{W}_{l} \) is the weight matrix of layer \( l \), \( \mathbf{X}_{l} \) is the input to layer \( l \), and \( \odot \) denotes element-wise multiplication. For example, Wanda uses a simple metric \( S_{ij} = |W_{ij}| \cdot \|X_{j}\|_2 \), achieving results comparable to methods involving the Hessian inverse \cite{Sun2023ASA_wanda}.

Instead of assigning a uniform sparsity ratio to all layers, our goal is to allocate different sparsity ratios while maintaining an overall model sparsity (e.g., 50\%). This allows critical layers to retain denser weights and less critical ones to be sparser, enhancing performance. The challenge is determining how to allocate sparsity ratios to different layers while keeping the overall sparsity constant.

\begin{algorithm}[t]
\caption{Sparsity Search Workflow} 
\label{alg1}
\begin{algorithmic}[1]
\REQUIRE Overall sparsity $\sigma_{\text{overall}}$, dataset $D$, max trials $T$, model $M$ 
\ENSURE Best sparsity profile $P_{\text{best}}$ 
\STATE $P_{\text{best}} \gets \emptyset$, $\text{PPL}_{\text{best}} \gets \infty$
\FOR{$t = 1$ to $T$}
    \STATE $P_t \gets$ Allocate Profile (TPE if $t > 1$)
    \IF{$\text{mean}(P_t) > \sigma_{\text{overall}}$}
        \STATE \textbf{continue} 
    \ENDIF
    \STATE Prune model $M$ with sparsity profile $P_t$
    \STATE $\text{PPL}_t \gets$ Evaluate pruned model on $D$
    \IF{$\text{PPL}_t < \text{PPL}_{\text{best}}$}
        \STATE $\text{PPL}_{\text{best}} \gets \text{PPL}_t$, $P_{\text{best}} \gets P_t$
    \ENDIF
\ENDFOR
\RETURN $P_{\text{best}}$
\end{algorithmic}
\end{algorithm}

One approach is to design a loss function as a proxy for model performance and apply gradient descent to optimize the sparsity masks. By incorporating sparsity masks into the reconstruction loss (Equation \ref{eq:recon_loss}), we introduce a regularization term to control overall sparsity:

\begin{equation}
\text{Loss} = \sum_{l} \| \mathbf{W}_{l}\mathbf{X}_{l} - (\mathbf{M}_{l} \odot \mathbf{W}_{l}) \mathbf{X}_{l} \|_2^2 + \lambda \left( \sum_{l} \|\mathbf{M}_{l}\|_0 - k \right)^2
\label{eq:total_loss}
\end{equation}

where \( \lambda \) is a regularization parameter, \( \|\mathbf{M}_{l}\|_0 \) is the number of non-zero elements in \( \mathbf{M}_{l} \), and \( k \) is the target number of retained weights.

However, this method has limitations:

\begin{itemize}
    \item The loss function must accurately reflect model performance and be convex.
    \item Balancing the reconstruction loss and regularization term is challenging and may not yield the desired overall sparsity.
\end{itemize}

\subsection{KV Cache Compression Allocation}

Uniform asymmetric quantization compresses the Key-Value (KV) Cache in LLMs by mapping high-precision values to lower-precision formats, reducing memory usage during inference. Given a high-precision tensor \( \mathbf{A} \in \mathbb{R}^{n \times d} \) and target bit-width \( b \), quantization is formulated as:

\begin{equation}
\text{Quant}_b(\mathbf{A})_{ij} = \text{round}\left( \frac{A_{ij} - \min(\mathbf{A})}{(\max(\mathbf{A}) - \min(\mathbf{A})) / {(2^b - 1)}} \right)
\end{equation}

This compresses \( \mathbf{A} \) into a lower bandwidth representation. Our sparsity search algorithm adapts to find optimal quantization bit-widths. Instead of uniformly applying 16-bit precision, we explore allocating different bit-widths (e.g., 6-bit, 8-bit) to various layers, maximizing compression while minimizing performance loss. Pruning and quantization can be performed independently or combined. By dynamically assigning bit-widths based on layer importance, critical layers retain higher precision, optimizing compression without significantly impacting performance.

\subsection{Layer-wise Search for Adaptive Allocation}

To overcome issues with the loss function approach, we adopt an iterative method to optimize layer-wise sparsity ratios, referring to each iteration as a \textbf{trial}. In each trial, we assign a sparsity ratio to each layer from a range around the overall sparsity (e.g., for 50\% overall sparsity, layers choose ratios from 45\% to 55\%). Our framework allows customizing ratios for different modules within a transformer layer (e.g., Q, K, V, O, MLPs). The collection of layer-specific ratios forms the sparsity profile for the trial. This allocation step is efficient and quick, not involving model loading or pruning. We ensure the average sparsity meets or exceeds the specified overall sparsity before proceeding.

Next, we prune the model based on the sparsity profile, ensuring each layer's sparsity matches the profile. We evaluate the pruned model's performance on Wikitext2, recording the perplexity (PPL). The goal is to minimize PPL over multiple trials. Using results from each trial, the TPE algorithm (detailed in the next section) allocates sparsity ratios for the next trial, optimizing the search space over random allocation. PPL is chosen as it reflects model performance and allows quick trial completion. Trials continue until PPL stabilizes or maximum trials are reached, concluding with the best-performing sparsity profile. Thanks to post-training pruning not requiring weight adjustments, a trial completes in 1-2 minutes (e.g., LLaVA 7B on A100). Typically, around 30 trials suffice to outperform uniform sparsity allocation.

\subsection{Tree-structured Parzen Estimator}

The TPE algorithm is an efficient Bayesian optimization method for hyperparameter search problems like our layer-wise sparsity allocation. TPE constructs a probabilistic model to explore the hyperparameter space more intelligently than grid or random search. It divides the hyperparameter space into two regions: one where hyperparameters perform well and one where they do not, modeling these separately using Parzen window estimators \cite{Watanabe2023Tree}. This allows informed decisions about the next hyperparameters to evaluate. The iterative process efficiently narrows the search space and converges to the optimal solution.

Advantages of using TPE include:

\begin{itemize}
    \item \textbf{Efficiency}: TPE selects hyperparameters likely to improve performance, reducing evaluations.
    \item \textbf{Flexibility}: It handles various hyperparameter types, including continuous, discrete, and conditional.
    \item \textbf{Adaptability}: TPE updates and optimizes its search strategy based on previous trial results.
\end{itemize}

\begin{table*}[t]
\caption{Comparison of our compressed LLaVA models with other large multimodal models on four benchmarks. 'PreTrain' and 'FineTune' indicate the dataset size used.}
\begin{center}
\begin{tabular}{|c|c|c|c|c|c|c|c|c|}
\hline
\textbf{Method} & \textbf{LLM} & \textbf{ImageSize} & \textbf{PreTrain} & \textbf{FineTune} & \textbf{VQAv2} & \textbf{SQA} & \textbf{TextVQA} & \textbf{POPE} \\
\hline
BLIP-2~\cite{li2023blip} & Vicuna-13B & $224^2$ & 129M & - & 41.00 & 61.00 & 42.50 & 85.30 \\
\hline
InstructBLIP~\cite{Dai2023InstructBlip} & Vicuna-7B & $224^2$ & 129M & 1.2M & - & 60.50 & 50.10 & - \\
\hline
InstructBLIP~\cite{Dai2023InstructBlip} & Vicuna-13B & $224^2$ & 129M & 1.2M & - & 63.10 & 50.70 & 78.90 \\
\hline
Shikra~\cite{chen2023shikra} & Vicuna-13B & $224^2$ & 600K & 5.5M & 77.40 & - & - & - \\
\hline
IDEFICS-9B~\cite{IDEFICS2023} & LLaMA-7B & $224^2$ & 353M & 1M & 50.90 & - & 25.90 & - \\
\hline
IDEFICS-80B~\cite{IDEFICS2023} & LLaMA-65B & $224^2$ & 353M & 1M & 60.00 & - & 30.90 & - \\
\hline
Qwen-VL~\cite{Bai2023QwenVL} & Qwen-7B & $448^2$ & 1.4B & 50M & 78.80 & 67.10 & 63.80 & - \\
\hline
Qwen-VL-Chat~\cite{Bai2023QwenVL} & Qwen-7B & $448^2$ & 1.4B & 50M & 78.20 & 68.20 & 61.50 & - \\
\hline
\multicolumn{9}{|c|}{\textbf{LLaVA-1.5~\cite{liu2023improvedllava}}} \\
\hline
LLaVA-1.5 & Vicuna-7B & $336^2$ & 558K & 665K & 78.50 & 66.80 & 58.20 & 85.90 \\
\hline
LLaVA-1.5 (Pruning) & Vicuna-7B & $336^2$ & 558K & 665K & 76.08 & 65.57 & 54.36 & 87.35 \\
\hline
LLaVA-1.5 (Pruning \& KV Cache Quant) & Vicuna-7B & $336^2$ & 558K & 665K & 75.90 & 64.91 & 54.14 & 87.42 \\
\hline
\multicolumn{9}{|c|}{\textbf{LLaVA-1.5~\cite{liu2023improvedllava}}} \\
\hline
LLaVA-1.5 & Vicuna-13B & $336^2$ & 558K & 665K & 80.00 & 74.94 & 61.30 & 85.90 \\
\hline
LLaVA-1.5 (Pruning) & Vicuna-13B & $336^2$ & 558K & 665K & 79.10 & 73.17 & 58.70 & 83.94 \\
\hline
LLaVA-1.5 (Pruning \& KV Cache Quant) & Vicuna-13B & $336^2$ & 558K & 665K & 79.08 & 72.95 & 58.51 & 84.11 \\
\hline
\multicolumn{9}{l}{'-' denotes that the model is unavailable for the current task.}
\end{tabular}
\label{table_1}
\end{center}
\end{table*}

\begin{table*}[t]
\caption{Benchmarks of the dense and pruned LLaVA-1.5 models (Vicuna-7B and Vicuna-13B) with different unstructured pruning methods. Bold indicates the best result of compressed models for each column.}
\begin{center}
\begin{tabular}{|c|c|c|c|c|c|c|}
\hline
\textbf{Method} & \textbf{Overall Sparsity} & \textbf{VQAv2} & \textbf{SQA} & \textbf{TextVQA} & \textbf{POPE} & \textbf{Wikitext PPL} \\
\hline
\multicolumn{7}{|c|}{\textbf{LLaVA-1.5 (Vicuna-7B)}} \\
\hline
Dense & 0\% & 78.50 & 66.80 & 58.20 & 85.90 & 6.40 \\
\hline
Magnitude & 50\% & 63.50 & 31.24 & 38.39 & 84.09 & 15.03 \\
\hline
SparseGPT & 50\% & 75.86 & 63.92 & 53.69 & 86.63 & 8.26 \\
\hline
Wanda & 50\% & 75.72 & 63.99 & 53.05 & 87.10 & 8.35 \\
\hline
Ours & 50\% & \textbf{76.08} & \textbf{65.57} & \textbf{54.36} & \textbf{87.35} & \textbf{8.00} \\
\hline
\multicolumn{7}{|c|}{\textbf{LLaVA-1.5 (Vicuna-13B)}} \\
\hline
Dense & 0\% & 80.00 & 74.94 & 61.30 & 85.90 & 5.61 \\
\hline
Magnitude & 50\% & 75.79 & 70.95 & 52.16 & 75.20 & 7.65 \\
\hline
SparseGPT & 50\% & 78.62 & 71.19 & 58.23 & \textbf{86.90} & 6.88 \\
\hline
Wanda & 50\% & 78.58 & 70.97 & 58.03 & 85.01 & 6.86 \\
\hline
Ours & 50\% & \textbf{79.10} & \textbf{73.17} & \textbf{58.70} & 83.94 & \textbf{6.66} \\
\hline
\end{tabular}
\label{table_2}
\end{center}
\end{table*}

\subsection{Design of Pruning Metric}

\noindent\textbf{Motivation} Traditional pruning methods often use simple metrics like weight magnitude, which may not accurately reflect unit importance. To balance computational efficiency and accurate weight importance representation, we introduce a novel pruning metric inspired by Wanda that addresses complexities in weight and activation distributions, enhancing pruning efficiency and robustness. Our proposed pruning metric, denoted as \( \mathcal{S}(w_{i,j}) \) for the weight \( w_{i,j} \) at position \( i,j \), is defined as:

\begin{equation}
\mathcal{S}(w_{i,j}) = \log\left(1 + \frac{|W_{i,j}|}{\|W_i\|_2} + \frac{|W_{i,j}|}{\|W_j\|_2}\right) \cdot \sqrt{\|X_j\|_2}
\label{eq:pruning_metric}
\end{equation}

Here, \( W_{i,j} \) is the weight at position \( i,j \); \( \|W_i\|_2 \) and \( \|W_j\|_2 \) are the L2 norms of the \( i \)-th row and \( j \)-th column, normalizing the weight's relative importance within its row and column. The logarithm enhances robustness by mitigating outliers, and \( \sqrt{\|X_j\|_2} \) balances magnitude and activation importance.
\noindent Advantages:
\begin{itemize}
    \item \textbf{High Computational Speed}: Our metric is computationally efficient, avoiding complex calculations like weight adjustments or Hessian inverses, allowing pruning of the LLaVA 7B model on an A100 GPU in 1-2 minutes. Efficient vectorized computation further accelerates the process.
    \item \textbf{Robustness}: Compared to simpler metrics like Wanda, our metric includes weight sorting in both row and column dimensions, L2 normalization, and a logarithmic function, enhancing robustness to extreme values and ensuring stable pruning results. By considering normalized weights and their distributions, it thoroughly assesses unit importance, capturing intricate weight and activation patterns.
\end{itemize}

\section{Experiments}

\subsection{Experimental Setup}

\noindent\textbf{Infrastructure}: Experiments were conducted on Ubuntu 20.04.3 LTS with CUDA 12.0, a 64-core CPU, 64\,GB RAM, and a single NVIDIA A100 GPU (80\,GB). A NVIDIA V100 GPU (32\,GB) suffices for running LLaVA 7B or 13B models. Required software packages are listed in the code repository.

\noindent\textbf{Models}: We applied our compression methods to LLaVA 1.5 models with 7B and 13B parameters. Additionally, we tested single-modality LLMs like LLaMA 1 and 2 (7B and 13B) to explore generalization capabilities.

\noindent\textbf{Benchmarks}: Evaluation was performed on visual question-answering and reasoning benchmarks including VQAv2~\cite{goyal2017vqav2}, ScienceQA~\cite{lu2022learn}, TextVQA~\cite{singh2019textvqa}, and the POPE hallucination benchmark~\cite{li2023pope}, following LLaVA's guidelines. For single-modality evaluation, we used Wikitext2~\cite{merity2016pointer_wikitext2} to measure perplexity. These benchmarks are open-source, easy to evaluate, and do not incur additional costs like GPT-4 API calls.

\noindent\textbf{Baselines}: For pruning, we compared against post-training methods Wanda and SparseGPT using their default unstructured pruning settings. We also included traditional magnitude pruning and the original dense models for reference. For quantization, results from various bandwidth choices were provided. Only one algorithm run was needed for these baselines.

\noindent\textbf{Our Method}: In pruning experiments, our framework applied unstructured pruning under various overall sparsity ratios. The key differences from Wanda and SparseGPT are: (1) our advanced pruning metric, and (2) adaptive layer-wise sparsity allocation based on the best sparsity profile found over 50 trials using Algorithm~\ref{alg1}. The sparsity ratios for each layer were chosen within $\pm5\%$ of the overall ratio, in $2.5\%$ increments. In quantization experiments, we applied uniform KV cache quantization to models pruned at 50\% overall sparsity. Under the constraint that half the layers used 8-bit bandwidth and the other half used 6-bit, our framework searched for the optimal mixed bandwidth allocation.

\subsection{Experimental Results}

% \begin{table}[tbp]
% \setlength{\tabcolsep}{1.5mm}
%   \centering
%     \begin{tabular}{ccccccc}
%     \toprule
%     \multirow{2}{*}{Model} & \multicolumn{2}{c}{LLaMA1} & \multicolumn{2}{c}{LLaMA2} & \multicolumn{2}{c}{LLaVA1.5} \\
%     \cmidrule(r){2-3} \cmidrule(r){4-5} \cmidrule(r){6-7}
%           & 7B   & 13B  & 7B   & 13B  & 7B   & 13B \\
%     \midrule
%     Dense & 5.68 & 5.09 & 5.12 & 4.57 & 6.40 & 5.61 \\
%     Magnitude & 17.29 & 20.21 & 14.90 & 6.37 & 15.03 & 7.65 \\
%     SparseGPT & 7.17 & 6.21 & 6.51 & 5.63 & 8.26 & 6.88 \\
%     Wanda & 7.26 & 6.15 & 6.46 & 5.56 & 8.35 & 6.86 \\
%     Ours  & \textbf{7.07} & \textbf{6.03} & \textbf{6.36} & \textbf{5.46} & \textbf{8.00} & \textbf{6.66} \\
%     \bottomrule
%     \end{tabular}%
% \caption{WikiText2 perplexity of unstructured pruned LLaMA and LLaVA series under the overall 50\% sparsity ratio. Bold indicates the best result of each column.}
% \label{table_3}%
% \end{table}%

\begin{table}[tbp]
\caption{WikiText2 perplexity of unstructured pruned LLaMA and LLaVA series under the overall 50\% sparsity ratio. Bold indicates the best result of compressed models.}
\begin{center}
\begin{tabular}{|c|c|c|c|c|c|c|}
\hline
\textbf{Model} & \multicolumn{2}{|c|}{\textbf{LLaMA1}} & \multicolumn{2}{|c|}{\textbf{LLaMA2}} & \multicolumn{2}{|c|}{\textbf{LLaVA1.5}} \\
\hline
 & \textbf{7B} & \textbf{13B} & \textbf{7B} & \textbf{13B} & \textbf{7B} & \textbf{13B} \\
\hline
Dense & 5.68 & 5.09 & 5.12 & 4.57 & 6.40 & 5.61 \\
\hline
Magnitude & 17.29 & 20.21 & 14.90 & 6.37 & 15.03 & 7.65 \\
\hline
SparseGPT & 7.17 & 6.21 & 6.51 & 5.63 & 8.26 & 6.88 \\
\hline
Wanda & 7.26 & 6.15 & 6.46 & 5.56 & 8.35 & 6.86 \\
\hline
Ours & \textbf{7.07} & \textbf{6.03} & \textbf{6.36} & \textbf{5.46} & \textbf{8.00} & \textbf{6.66} \\
\hline
\end{tabular}
\label{table_3}
\end{center}
\end{table}

% \begin{table}[t]
% \setlength{\tabcolsep}{1.5mm}
% \centering
% \begin{tabular}{ccccccc}
% \toprule
% Bandwidth & TextVQA & SQA & VQAv2 & POPE & PPL \\
% \midrule
% \multicolumn{6}{c}{\textbf{LLaVA-1.5 (Vicuna-7B)}} \\
% 16 & 54.36 & 65.57 & 76.08 & 87.35 & 8.00 \\
% 8 & 54.21 & 65.22 & 76.02 & 87.53 & 8.07 \\
% 6 & 49.05 & 11.79 & 74.06 & 86.96 & 32.63 \\
% Opposite & 49.45 & 51.87 & 74.25 & 86.56 & 31.82 \\
% Ours & 54.14 & 64.91 & 75.90 & 87.42 & 8.24 \\
% \midrule
% \multicolumn{6}{c}{\textbf{LLaVA-1.5 (Vicuna-13B)}} \\
%  16 & 58.70 & 73.17 & 79.10 & 83.94 & 6.66 \\
%  8 & 58.42 & 72.84 & 79.10 & 83.99 & 6.71 \\
%  6 & 55.10 & 66.14 & 77.79 & 83.89 & 8.73 \\
%  Opposite & 55.14 & 66.12 & 77.92 & 83.94 & 8.33 \\
%  Ours & 58.51 & 72.95 & 79.08 & 84.11 & 6.96 \\
% \bottomrule
% \end{tabular}
% \caption{Performance of pruned models under unstructured overall 50\% sparsity with different KV cache uniform quantization bandwidths.} 
% \label{table_quant}
% \end{table}

\begin{table}[t]
\caption{Performance of pruned models under unstructured overall 50\% sparsity with different KV cache uniform quantization bandwidths.}
\begin{center}
\begin{tabular}{|c|c|c|c|c|c|}
\hline
\textbf{Bandwidth} & \textbf{TextVQA} & \textbf{SQA} & \textbf{VQAv2} & \textbf{POPE} & \textbf{PPL} \\
\hline
\multicolumn{6}{|c|}{\textbf{LLaVA-1.5 (Vicuna-7B)}} \\
\hline
16 & 54.36 & 65.57 & 76.08 & 87.35 & 8.00 \\
\hline
8 & 54.21 & 65.22 & 76.02 & 87.53 & 8.07 \\
\hline
6 & 49.05 & 11.79 & 74.06 & 86.96 & 32.63 \\
\hline
Opposite & 49.45 & 51.87 & 74.25 & 86.56 & 31.82 \\
\hline
Ours & 54.14 & 64.91 & 75.90 & 87.42 & 8.24 \\
\hline
\multicolumn{6}{|c|}{\textbf{LLaVA-1.5 (Vicuna-13B)}} \\
\hline
16 & 58.70 & 73.17 & 79.10 & 83.94 & 6.66 \\
\hline
8 & 58.42 & 72.84 & 79.10 & 83.99 & 6.71 \\
\hline
6 & 55.10 & 66.14 & 77.79 & 83.89 & 8.73 \\
\hline
Opposite & 55.14 & 66.12 & 77.92 & 83.94 & 8.33 \\
\hline
Ours & 58.51 & 72.95 & 79.08 & 84.11 & 6.96 \\
\hline
\end{tabular}
\label{table_quant}
\end{center}
\end{table}

\noindent\textbf{Comparison to Conventional Large Multimodal Models} Table~\ref{table_1} compares the performance of our compressed LLaVA-1.5 models (7B and 13B) with other state-of-the-art multimodal models on four benchmarks: VQAv2, SQA, TextVQA, and POPE. In the table, 'with pruning' indicates unstructured pruning using our method at a 50\% overall sparsity ratio, and 'with pruning + quant' adds our quantization method, allocating half of the layers to 6-bit and the other half to 8-bit bandwidths. Despite pruning 50\% of the weights and applying quantization, our framework enables LLaVA-1.5 models to maintain competitive performance with uncompressed models.

On VQAv2, the pruned LLaVA-1.5 with Vicuna-13B scores 79.10, comparable to unpruned models like Qwen-VL (78.80) and slightly below the original LLaVA-1.5 (80.00). For SQA, our pruned model scores 73.17, outperforming models like BLIP-2 (61.00) and InstructBLIP (63.10), and is only slightly lower than the unpruned LLaVA-1.5 (74.94). In TextVQA, the pruned model achieves 58.70, maintaining strong performance despite significant weight reduction and outperforming models like InstructBLIP (50.70). Quantization introduces a slight performance loss, but our method's layer-wise bandwidth allocation mitigates this effect.

\noindent\textbf{Comparison to Conventional Pruning Methods} Table~\ref{table_2} shows unstructured pruning results for LLaVA-1.5 models with Vicuna-7B and Vicuna-13B backbones. We evaluated different pruning methods—Magnitude, SparseGPT, Wanda, and ours—across multimodal benchmarks (VQAv2, SQA, TextVQA, POPE) and a single-modality benchmark (WikiText), aiming for minimal performance loss at a high overall sparsity of 50\%.

In LLaVA-1.5 (Vicuna-7B), our method outperforms others on all benchmarks, achieving scores closest to the dense model. Traditional methods like Magnitude show significant performance drops, while SparseGPT and Wanda perform better but do not match our method's balance. The trend is consistent with LLaVA-1.5 (Vicuna-13B), indicating our method's effectiveness in maintaining model quality. Notably, in the POPE benchmark, the pruned model sometimes surpasses the dense model due to averaging three F1 scores.

To test generalization on single-modality LLMs, we evaluated pruned LLaMA 1 and 2 series in Table~\ref{table_3}. Results align with multimodal benchmarks, with our method achieving better overall performance.

Our method's advantage lies in individually optimizing layer-wise sparsity ratios within 45\% to 55\% while ensuring an overall sparsity of 50\%. This tailored approach, combined with our robust pruning metric, retains critical information in key layers, minimizing the impact of pruning on performance.

\begin{table}[tbp]
\caption{WikiText perplexity of unstructured pruned LLaVA-1.5 (Vicuna-7B and 13B) models with various sparsity ratios. Bold indicates the best result of each column.}
\begin{center}
\begin{tabular}{|c|c|c|c|c|c|c|}
\hline
\textbf{Model} & \textbf{Methods} & \textbf{10\%} & \textbf{20\%} & \textbf{30\%} & \textbf{40\%} & \textbf{50\%} \\
\hline
\multirow{4}{*}{7B} & Magnitude & \textbf{6.40} & 6.68 & 7.40 & 9.03 & 15.03 \\
\cline{2-7}
 & SparseGPT & 6.50 & 6.75 & 6.95 & 7.37 & 8.26 \\
\cline{2-7}
 & Wanda & 6.47 & 6.66 & 6.88 & 7.26 & 8.35 \\
\cline{2-7}
 & Ours & 6.41 & \textbf{6.47} & \textbf{6.64} & \textbf{7.03} & \textbf{8.00} \\
\hline
\multirow{4}{*}{13B} & Magnitude & 5.62 & 5.68 & 5.85 & 6.34 & 7.65 \\
\cline{2-7}
 & SparseGPT & 5.71 & 5.82 & 5.99 & 6.22 & 6.88 \\
\cline{2-7}
 & Wanda & 5.73 & 5.85 & 5.98 & 6.20 & 6.86 \\
\cline{2-7}
 & Ours & \textbf{5.62} & \textbf{5.68} & \textbf{5.84} & \textbf{6.07} & \textbf{6.66} \\
\hline
\end{tabular}
\label{table_various_ratios}
\end{center}
\end{table}

\noindent\textbf{Quantization Bandwidth Allocation}
Table~\ref{table_quant} examines how layer-wise bandwidth allocation affects model performance and seeks optimal allocation patterns. Unlike pruning, where sparsity impacts performance gradually, quantization bandwidth has abrupt effects: 8-bit quantization maintains performance close to the 16-bit baseline, while 6-bit leads to significant degradation, and 4-bit causes model failure. Therefore, we focused on optimizing allocations between 6-bit and 8-bit quantization to minimize performance loss. To prevent the bandwidth search from favoring higher bit-widths (due to the lack of an overall constraint like sparsity), we constrained the number of layers using 8-bit and 6-bit quantization to be equal, allowing us to explore optimized allocation schemes. Our bandwidth search results are presented in the 'Ours' row; the 'Opposite' row shows the reverse allocation for comparison.

The results reveal a significant performance gap between 'Ours' and 'Opposite', highlighting the importance of proper bandwidth allocation. Even with the same overall compression level, correct bandwidth distribution minimizes performance loss, while poor allocation severely degrades model quality. This demonstrates our framework's effectiveness in finding optimal bandwidth allocations that preserve model performance under aggressive quantization.

\noindent\textbf{Insights on Allocation}
Our experiments yield the following key insights:
\begin{itemize}
    \item \textbf{Pruning}: The framework assigns lower sparsity (denser weights) to early and late layers, and higher sparsity to middle layers, forming a U-shaped pattern. This indicates the importance of preserving more weights in the initial and final layers.
    \item \textbf{Quantization}: In contrast, higher precision (8-bit) is allocated to earlier layers, and lower precision (6-bit) to later layers. This suggests early layers require more precision, while later layers can function effectively with less.
\end{itemize}
These patterns highlight the distinct needs of different model sections in pruning and quantization, offering valuable guidance for optimization.

% \begin{table}[tbp]
%   \centering
%   \caption{Ablation Study for Search Strategies in LLaVA-1.5(Vicuna-7B) for the 50\% sparsity ratio.}
%     \begin{tabular}{ccccc}
%     \toprule
%    Strategies & Sparsity & WIKI PPL & SQA   & VQA \\
%     \midrule
%     Search ntrial=30 & 50\%  &         8.03  &       65.72  &       54.06  \\
%     Search ntrial=50 & 50\%  &         8.09  &       64.82  &       54.21  \\
%     Search ntrail=100 & 50\%  &         8.03  &       64.25  &       53.68  \\
%     \midrule
%     Search random  & 50\%  & 8.92  & 63.76 & 53.33 \\
%     Search grid  & 50\%  & 8.45  & 63.88 &       53.22  \\
%     Search TPE  & 50\%  &         8.03  &       65.72  &       54.06  \\
%     \bottomrule
%     \end{tabular}%
%   \label{tab:Search}%
% \end{table}%

\subsection{Ablation Studies}
% \noindent\textbf{Analysis for different metrics in our method.}
% Table \ref{tab:metric} compares our proposed pruning metric in formula \ref{eq:pruning_metric} with other similar metrics, using Wanda’s metric as the baseline. The experiment explores the impact of different normalization approaches on model performance. The results show that applying normalization across both row and column dimensions improves performance, with our method, which also incorporates a log1p function for added robustness, achieving the best results across all benchmarks.

\noindent\textbf{Comparison under different sparse ratios.} In addition to fixing the overall high sparsity ratio at 50\%, we also explored the performance of pruned models under various sparsity ratios ranging from 10\% to 50\%, evaluated by Wikitext perplexity in Table \ref{table_various_ratios}. The result showed that our method maintained a satisfying performance across the wide range of sparsity ratios. 

% \textbf{Analysis for Search Strategies in our method}
% The TPEoptimization used in our method surpasses random and grid search strategies, efficiently identifying optimal configurations with fewer trials. This effectiveness in navigating the parameter space leads to superior performance in WikiText perplexity, SQA, and VQA, emphasizing the importance of a well-designed search strategy in achieving high-quality pruning outcomes.

\section{Conclusion}
In this work, we present a novel framework for efficient compression of LMMs. At the core of this framework is an innovative adaptive search algorithm that uses the Tree-structured Parzen estimator to optimize pruning ratios and quantization bandwidth across different layers, with the LMM's performance as the optimization objective. Extensive evaluations on benchmark datasets, demonstrate that our method consistently outperforms existing state-of-the-art techniques. The framework provides a robust and systematic approach to compressing large multimodal models, enabling optimized performance under constrained computational resources and compression budgets, paving the way for efficient LMM deployment on edge devices. Our future work will explore extending this framework to support additional compression techniques, such as adaptive quantization and mixed-precision approaches, further enhancing the efficiency and flexibility of LMM compression.

\bibliographystyle{IEEEbib}
\bibliography{main}

\end{document}